\documentclass{article} 
\usepackage{iclr2027_conference,times}

\usepackage{amsmath,amsfonts,bm}

\def\eqref#1{equation~\ref{#1}}

\def\1{\bm{1}}

\DeclareMathAlphabet{\mathsfit}{\encodingdefault}{\sfdefault}{m}{sl}
\SetMathAlphabet{\mathsfit}{bold}{\encodingdefault}{\sfdefault}{bx}{n}

\iclrfinalcopy

\usepackage{hyperref}
\usepackage{url}
\usepackage{booktabs}
\usepackage[table]{xcolor}
\usepackage{graphicx}
\usepackage{multirow}
\usepackage{adjustbox}

\title{IDOL: Inverse-Dynamics-Guided Future Prediction for End-to-End Autonomous Driving}

\author{Chenghao Zhang \quad Timin Li \quad Dongmei Li \\
Department of Electronic Engineering, Tsinghua University \\
\texttt{\{zhangch24, litm22\}@mails.tsinghua.edu.cn}
\quad
\texttt{lidmei@tsinghua.edu.cn}
}

\begin{document}

\maketitle

\lhead{}

\vspace{-2.5em}
\begin{center}
\textbf{Code:} \url{https://github.com/Waterloo-ins/IDOL}
\end{center}

\begin{abstract}
End-to-end autonomous driving has emerged as a compelling paradigm for learning planning directly from sensor observations, while recent world-model-based approaches further enrich this paradigm by enabling explicit reasoning about how the scene may evolve in the future. Yet future prediction alone does not guarantee better planning unless the predicted evolution can be converted into planning-relevant trajectory updates. Many current methods still forecast future scene states without explicitly decoding the motion implications hidden in state transitions. As a result, future reasoning often remains descriptively useful but only weakly coupled to executable motion generation. To address this limitation, we propose \textbf{IDOL}, an inverse-dynamics-guided future prediction framework for world-model-based end-to-end planning in latent BEV space, where inverse dynamics serves as the key bridge between future prediction and trajectory optimization. IDOL first predicts multiple future latent scene states with a BEV world model, then applies an inverse dynamics model to adjacent latent futures to decode transition-aware trajectory features and recover planning-relevant motion deltas that explain how the latent world evolves over time. These transition features refine the candidate-specific planning
queries through spatial cross-attention and global modulation.
A lightweight closed-loop refinement module feeds the refined queries
back into candidate-conditioned latent rollouts, enabling an additional
round of future-informed query refinement before trajectory decoding. By introducing inverse dynamics into latent future reasoning, IDOL tightens the coupling between world modeling and planning. Extensive experiments on the NAVSIM v1 and NAVSIM v2 benchmarks show that IDOL achieves state-of-the-art performance among comparable methods. 
\end{abstract}

\section{Introduction}

Autonomous driving~\citep{jiang2023vad,chen2024vadv2,sun2025sparsedrive,li2024hydra,liu2025bridgedrive,guo2026flowad,huang2026occdriver,tang2025plan} has advanced significantly in recent years, driven by a paradigm shift from modular perception-prediction-planning pipelines to unified end-to-end learning frameworks. Recent advances have further expanded this paradigm toward future-aware, world-model-guided planning~\citep{zhang2026resworld,xia2025drivelaw}, scalable sim-real training~\citep{tian2025simscale}, and video-centric pretraining~\citep{strong2026learning}, leading to markedly stronger robustness and planning performance on realistic benchmarks such as nuScenes~\citep{caesar2020nuscenes} and NAVSIM~\citep{dauner2024navsim}. At the same time, multimodal and generative formulations~\citep{xu2025wam,liu2025guideflow,li2025discrete,zhao2025diffe2e}  have shown that planning quality can benefit substantially from moving beyond deterministic single-pass regression.More recently,  VLA-based systems~\citep{li2026sgdrive,jiang2025diffvla,luo2025adathinkdrive} extended the scope of future-aware decision making by injecting language-grounded reasoning, action abstraction, and reflective post-training into driving systems.

\begin{figure}[t]
    \centering
    \includegraphics[width=\linewidth]{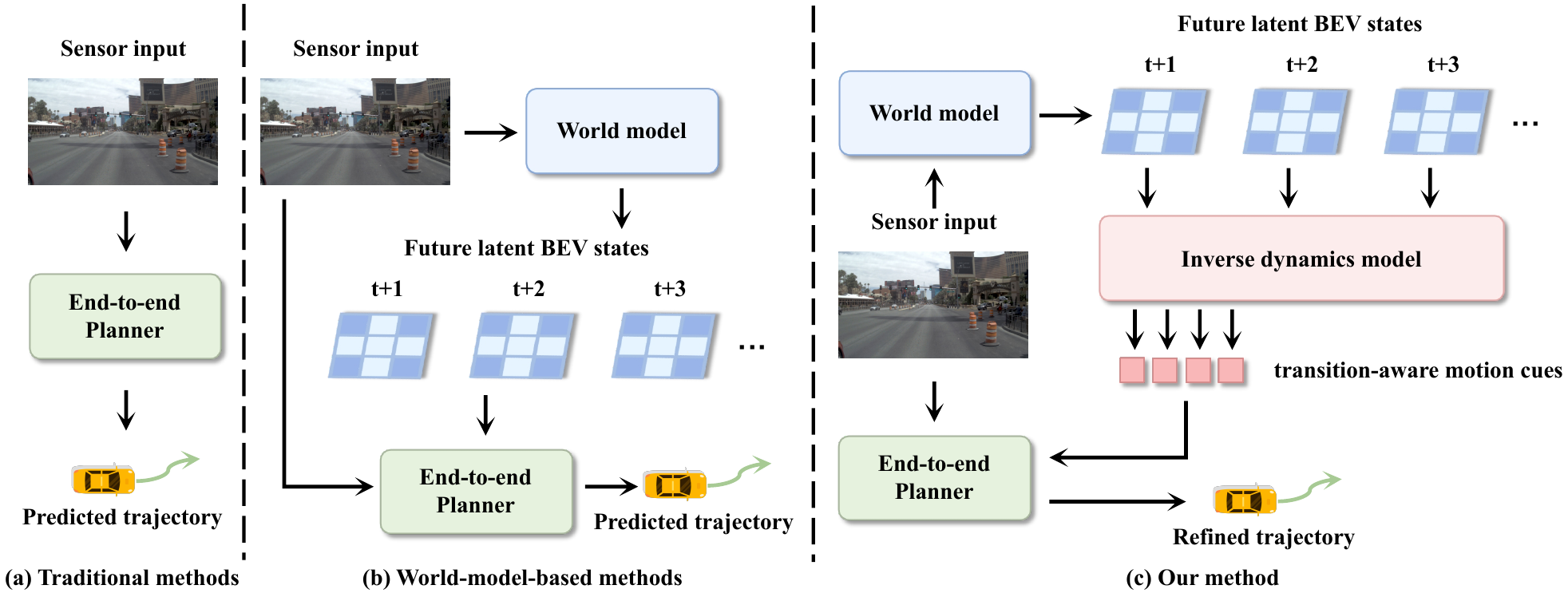}
    \caption{Comparison of three planning paradigms. (a) Traditional methods directly map sensor inputs to a predicted trajectory. (b) World-model-based methods additionally predict future latent BEV states to support planning, but these predicted futures are only loosely connected to motion generation. (c) In contrast, IDOL further applies an inverse dynamics model to adjacent future latent BEV states to extract transition-aware motion cues for trajectory refinement, making future prediction more actionable for planning.}
    \label{fig:figure1}
\end{figure}

For trajectory planning, traditional end-to-end autonomous driving methods~\citep{tang2026causalvad,wang2026meanfuser} typically rely on the current state, which constitutes a form of passive planning, as illustrated in Figure~\ref{fig:figure1}(a). Subsequently, many approaches~\citep{jiang2025wpt,kim2026safedrive} have introduced world models into end-to-end planning, adopting the predict-decide paradigm to make decision-making more foresighted, as illustrated in Figure~\ref{fig:figure1}(b).

Yet a fundamental bottleneck remains: even with future prediction, current methods~\citep{li2025end,zhang2025future} still lack a clear and interpretable bridge between forecasting scene evolution and using it to produce dynamically meaningful trajectory updates. Predicted futures may therefore be informative but insufficiently actionable: the planner can anticipate what may happen next, yet lacks a principled mechanism to revise the current trajectory so that it remains executable, dynamically coherent, and robust over long horizons. This limitation is especially critical in safety-sensitive driving, where future reasoning must ultimately support motion generation rather than passive forecasting. Our key insight is that \emph{the missing signal lies not only in the predicted future states themselves, but more importantly in the transitions between them}. As illustrated in Figure~\ref{fig:figure1}(c), inverse dynamics naturally decodes this signal. While forward world modeling predicts how the scene evolves under a motion hypothesis, inverse dynamics asks the complementary question of \emph{what motion-relevant change is implied by the transition from one state to the next}. In latent Bird's Eye View space, adjacent predicted futures capture the expected evolution of the ego vehicle and surrounding scene under a planning hypothesis, and their transitions therefore contain recoverable motion semantics reflecting trajectory adjustments consistent with that evolution. Rather than treating future latent states as static planning features, we decode adjacent latent transitions with an inverse dynamics model to extract transition-aware trajectory features and infer motion deltas for planning optimization. In this way, inverse dynamics explicitly bridges world modeling and trajectory generation, transforming future prediction from \emph{passive scene forecasting} into \emph{planning-aware transition reasoning}.

Based on this perspective, we propose \textbf{IDOL}, a world-model-based end-to-end planner operating in latent BEV space that tightly couples future reasoning with trajectory optimization. IDOL first predicts multiple future latent scene states with a BEV world model, providing anticipated scene evolution beyond the current observation. It then applies an inverse dynamics model to adjacent latent futures to infer motion-aware transition features and planning-relevant trajectory deltas that explain the predicted evolution. These signals are aggregated and injected into the planner to optimize the anchor trajectory, allowing motion updates to follow the dynamics implied by future state transitions rather than future context alone. A lightweight closed-loop refinement module further re-evaluates the updated trajectory to improve long-horizon consistency.

In summary, our contributions are as follows:

\noindent\textbf{i.} We identify a fundamental gap in recent end-to-end and world-model-based planners: current methods still lack a sufficiently explicit and interpretable mechanism for converting predicted future evolution into planning-relevant trajectory updates, which limits the actionability of future reasoning for executable motion generation.

\noindent\textbf{ii.} We propose IDOL, a world-model-based end-to-end planner in latent BEV space that introduces inverse dynamics into latent future reasoning, thereby establishing an explicit bridge between world modeling and executable motion generation. We further design a lightweight closed-loop refinement module to improve long-horizon consistency through another round of future-aware reasoning.

\noindent\textbf{iii.} We conduct extensive experiments on both NAVSIM v1 and NAVSIM v2, including comprehensive comparisons and analysis, and show that our method consistently achieves state-of-the-art performance among comparable methods. These results validate the effectiveness of our design in improving planning quality across different benchmark settings.

\section{Related work}
\label{gen_inst}

\subsection{End-to-end autonomous driving}

End-to-end autonomous driving~\citep{zhang2025future,li2025hydra,feng2025rap,li2024navigation,luo2025adathinkdrive,chen2024end,feng2025artemis,shang2026drivedpo} directly maps sensor observations to waypoints, trajectories, or low-level controls, reducing error accumulation and hand-crafted interfaces in modular pipelines. Early methods~\citep{chitta2022transfuser} largely relied on imitation learning, while planning-oriented formulations~\citep{guo2025ipad,wozniak2025prix,yao2026drivesuprim,hwang2024emma,huang2025gen,sun2025minddrive,li2025imagidrive,sun2026sparsedrivev2} later became dominant. TransFuser~\citep{chitta2022transfuser} established a strong transformer-based sensor-fusion baseline, while UniAD~\citep{hu2023planning} unified perception and prediction around downstream planning. DriveMamba~\citep{su2026drivemamba} improved efficiency and scalability with a task-centric state-space architecture, and LEAD~\citep{nguyen2025lead} revisited imitation learning through learner-expert asymmetry. DiffusionDrive~\citep{liao2025diffusiondrive} and GoalFlow~\citep{xing2025goalflow} introduced generative planning via truncated diffusion and flow matching to better model multimodal trajectories. More recently, VLA-based end-to-end driving~\citep{zheng2025driveagent,rawal2026nord,xu2025wam,wang2026vggdrive} has extended this direction toward language-grounded action generation and post-training alignment. DriveVLA-W0~\citep{li2025drivevla} incorporated world-model objectives to provide dense supervision and improve data scaling, while AutoDrive-$R^2$~\citep{yuan2026autodrive} introduced chain-of-thought reasoning, self-reflection, and GRPO-based post-training for physically plausible trajectory generation. Despite these advances, most end-to-end planners still couple planning mainly to the current scene or output-level multimodality, leaving the connection between future state transitions and action generation largely implicit.

\subsection{World model in autonomous driving}

World models in autonomous driving~\citep{chen2025drivinggpt,wang2024drivedreamer,li2024enhancing,jiang2025wpt} predict scene dynamics for forecasting, planning, simulation, and long-horizon decisions. Dreamer~\citep{hafner2019dream,hafner2020mastering,hafner2023mastering} introduced latent imagined rollouts for long-horizon planning. GAIA-1~\citep{hu2023gaia} formulated driving world modeling as action-conditioned multimodal sequence generation, while Vista~\citep{gao2024vista} emphasized environmental diversity and controllability. WoTE~\citep{li2025end} evaluates trajectory candidates through predicted BEV states. SeerDrive~\citep{zhang2025future} couples scene evolution and planning through iterative bidirectional refinement. DriveLaW~\citep{xia2025drivelaw} unifies video generation and motion planning in shared latents that support closed-loop planning.

Despite this progress, the key challenge is how predicted futures shape decisions. Although recent methods~\citep{wang2024driving,zheng2025world4drive,zhang2025future} jointly train world modeling and planning, the connection between future scene dynamics and planned motion remains insufficiently explained. We address this gap by making latent BEV transitions action-relevant, constraining planning through both anticipated scene evolution and its implied control signals.

\subsection{Inverse dynamics models}

Inverse dynamics models~\citep{hollerbach1980recursive,otten2003inverse,koopman1995inverse,angeles1989algorithm} originate from physics, control, and reinforcement learning, where the goal is to infer the action or control input that caused a transition between two states. This viewpoint is useful because it makes action semantics explicit: an inverse dynamics model can identify which aspects of a transition are action-relevant, recover latent controls when actions are unavailable or noisy, and encourage representations that preserve controllable structure rather than merely appearance. Recent machine learning work~\citep{tian2024predictive} has employed inverse dynamics for model-based control, representation learning, and general sequential decision-making.

\begin{figure}[t]
    \centering
    \includegraphics[width=\linewidth]{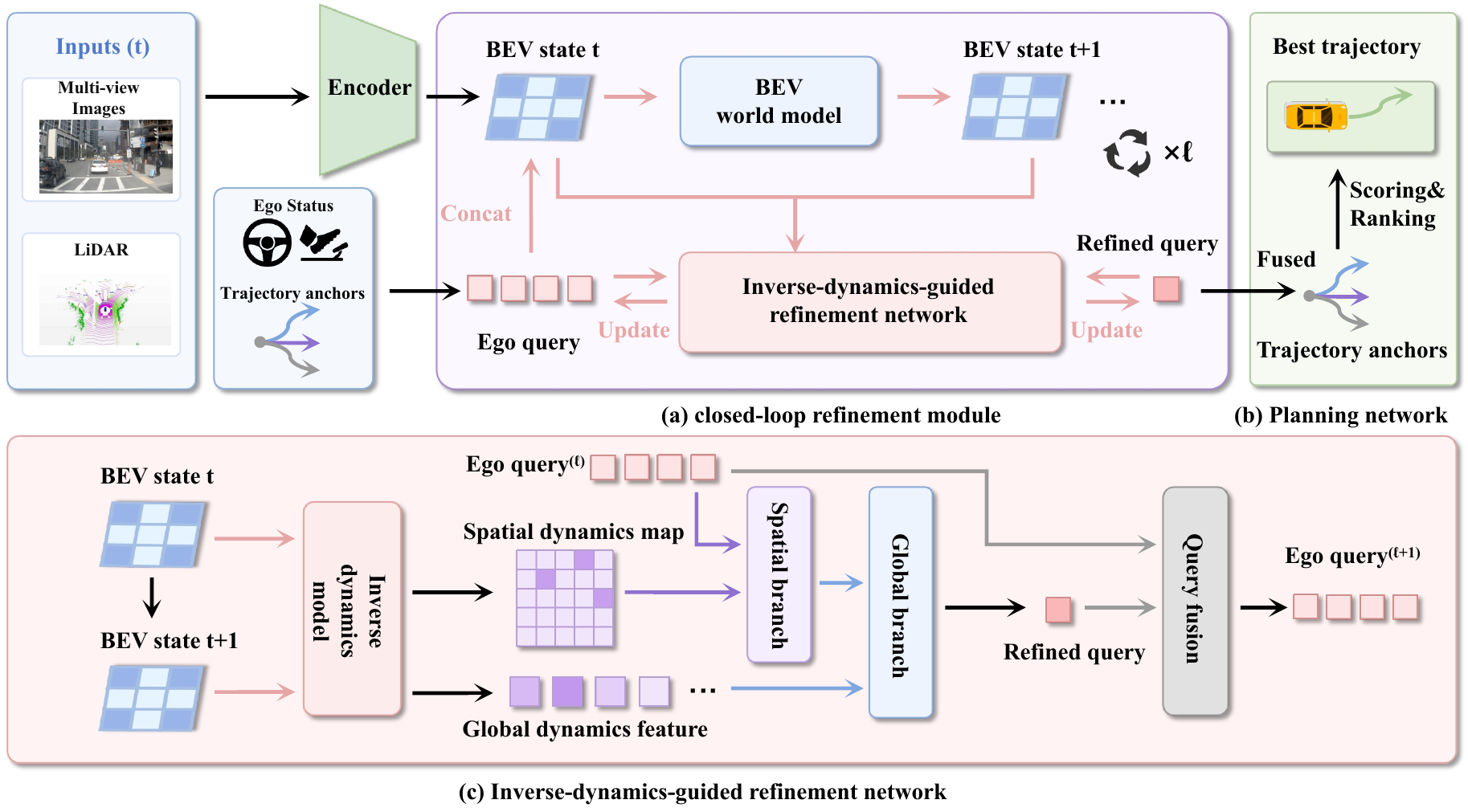}
    \caption{Overview of the proposed IDOL framework. (a) The closed-loop refinement module rolls out future latent BEV states with the BEV world model and refines the planning query through inverse-dynamics feedback. (b) The planning network fuses the refined query with trajectory anchors, predicts candidate trajectories, and selects the best one by scoring and ranking. (c) The inverse-dynamics-guided refinement network decodes adjacent future BEV states into spatial and global dynamics cues, which are fused to update the ego query.}
    \label{fig:figure2}
\end{figure}

In autonomous driving, inverse dynamics remains comparatively underexplored and is mostly used as an auxiliary component. ReSim~\citep{yang2025resim} employs an inverse dynamics model to convert predicted videos into executable trajectories and assess controllability, while FutureSightDrive~\citep{zeng2025futuresightdrive} formulates planning as recovering actions from current observations and imagined future scenes. These works suggest that inverse dynamics can bridge predicted future states and executable actions. However, existing driving world models typically focus on latent future forecasting, trajectory evaluation, or reward-guided selection, without explicitly learning transition-to-action mappings between adjacent predicted states. Our method addresses this gap by introducing an inverse dynamics module over predicted latent BEV states, explicitly coupling state evolution with planning and aligning trajectory generation with action-consistent future transitions.

\section{Methodology}
\label{headings}

\subsection{Preliminary}

\paragraph{End-to-end autonomous driving.}
End-to-end autonomous driving formulates trajectory planning as learning a direct mapping from raw sensory observations and ego-state inputs to future ego motion, without relying on a manually designed modular stack. Recent world-model-based methods further improve this paradigm by explicitly modeling future scene evolution and using the predicted futures to guide planning.

\paragraph{Inverse dynamics model.}
Inverse dynamics model infers the latent action or motion representation that explains the transition between two adjacent states. In its general form, given two consecutive states $\xi_t$ and $\xi_{t+1}$, inverse dynamics seeks a compact transition descriptor
\begin{equation}
d_t = f_{\mathrm{idm}}(\xi_t,\xi_{t+1}),
\end{equation}
which captures the motion-consistent change from $\xi_t$ to $\xi_{t+1}$. This formulation is particularly suitable for autonomous driving, since adjacent future states naturally encode how ego motion interacts with scene evolution. In our setting, the two states are imagined BEV latent states produced by the world model. Therefore, inverse dynamics provides a natural bridge from \emph{future state transition} to \emph{trajectory-related motion features}, making it well suited for refining trajectories from consecutive imagined BEV frames.

\subsection{Scene representation}

As shown in Figure~\ref{fig:figure2}, we encode the current scene into a latent BEV representation $B$ using a TransFuser-style image-LiDAR backbone~\citep{chitta2022transfuser}. We adopt an offline trajectory-anchor vocabulary $\mathcal{A}=\{\tau_k\}_{k=1}^{K}$, where $\tau_k\in\mathbb{R}^{S\times 3}$ represents a candidate future motion~\citep{chen2024vadv2,li2024hydra}. Each anchor feature $a_k$ is concatenated with the ego-state feature $z$ to initialize a candidate-specific planning query:
\begin{equation}
e_{k,\mathrm{init}}=\phi([z,a_k]),
\qquad k=1,\ldots,K,
\end{equation}
which guides subsequent future imagination and trajectory refinement.

\subsection{Latent BEV world model}

For each trajectory candidate $k$ and refinement iteration $\ell$, we initialize the latent rollout with
$q_{k,\ell}^{(0)}=e_{k,\mathrm{in}}^{(\ell)}$ and
$B_{k,\ell}^{(0)}=B$.
At each step $u$, the ego query is injected at the corresponding future location of anchor $\tau_k$, and a transformer-based world model predicts the next query and latent BEV state:
\begin{align}
B_{k,\ell,\mathrm{cond}}^{(u)}
&=
\mathcal{I}\!\left(
B_{k,\ell}^{(u)},q_{k,\ell}^{(u)};\tau_k
\right),\\
\left[q_{k,\ell}^{(u+1)},B_{k,\ell}^{(u+1)}\right]
&=
\mathrm{BEVWorldModel}\!\left(
\left[q_{k,\ell}^{(u)},B_{k,\ell,\mathrm{cond}}^{(u)}\right]
\right),
\end{align}
where $\mathcal{I}$ denotes query injection.
Repeating this process with the same anchor yields the candidate-specific rollout
\begin{equation}
\mathcal{R}_{k}^{(\ell)}
=
\left\{
\left(q_{k,\ell}^{(u)},B_{k,\ell,\mathrm{cond}}^{(u)}\right)
\right\}_{u=0}^{U},
\end{equation}
whose adjacent candidate-conditioned BEV states are subsequently decoded by the inverse dynamics model.

\subsection{Inverse-dynamics-guided refinement network}

\subsubsection{Inverse dynamics model}

While the world model predicts future latent scene evolution, planning still requires a mechanism that translates the imagined transitions into motion-relevant feedback. To bridge this gap, we introduce an inverse dynamics model (IDM).

For each pair of adjacent candidate-conditioned latent BEV states within the $\ell$-th rollout, the IDM produces two complementary transition representations:
\begin{equation}
\left(
S_{k,\ell}^{(u)},
g_{k,\ell}^{(u)}
\right)
=
\mathrm{IDM}\!\left(
B_{k,\ell,\mathrm{cond}}^{(u)},
B_{k,\ell,\mathrm{cond}}^{(u+1)}
\right),
\end{equation}
where
$S_{k,\ell}^{(u)}\in\mathbb{R}^{H\times W\times C}$
is a spatial dynamics map that preserves position-wise motion variations across the latent BEV feature map, and
$g_{k,\ell}^{(u)}\in\mathbb{R}^{C}$
is a pooled global dynamics feature that summarizes the overall transition.

When multiple rollout transitions are available, we aggregate the transition representations over the rollout horizon:
\begin{equation}
S_k^{(\ell)}
=
\frac{1}{U}
\sum_{u=0}^{U-1}
S_{k,\ell}^{(u)},
\qquad
g_k^{(\ell)}
=
\frac{1}{U}
\sum_{u=0}^{U-1}
g_{k,\ell}^{(u)}.
\end{equation}

The key idea is that future transitions should capture not only \emph{what} changes globally, but also \emph{where} critical changes occur. A pooled inverse-dynamics feature alone is often too coarse, since only part of the spatial variation is relevant to the current trajectory hypothesis. The spatial representation preserves localized transition evidence, while the global representation summarizes overall evolution. Together, they bridge latent future prediction and motion refinement at the transition level.

\subsubsection{Closed-loop query refinement}

We refine the planning query by combining the spatial and global inverse-dynamics cues through a dual-branch mechanism.

For the first closed-loop iteration, the refinement process is initialized by the anchor-conditioned query:
\begin{equation}
e_{k,\mathrm{in}}^{(0)}
=
e_{k,\mathrm{init}}.
\end{equation}

Given the transition representations extracted from the corresponding latent rollout, the spatial branch first allows the current planning query to retrieve candidate-relevant local motion evidence through cross-attention:
\begin{equation}
e_{k,\mathrm{spa}}^{(\ell)}
=
\mathrm{LN}\!\left(
e_{k,\mathrm{in}}^{(\ell)}
+
\mathrm{CrossAttn}\!\left(
e_{k,\mathrm{in}}^{(\ell)},
S_k^{(\ell)},
S_k^{(\ell)}
\right)
\right).
\end{equation}
Here, $e_{k,\mathrm{spa}}^{(\ell)}$ denotes the query after spatial transition-aware refinement.

The global dynamics representation is subsequently used to calibrate the spatially refined query through an MLN~\citep{wang2023exploring}:
\begin{equation}
e_{k,\mathrm{ref}}^{(\ell)}
=
\mathrm{MLN}_{\mathrm{idm}}\!\left(
e_{k,\mathrm{spa}}^{(\ell)},
g_k^{(\ell)}
\right).
\end{equation}
This dual-branch design assigns complementary roles to the two inverse-dynamics representations: the spatial branch determines \emph{where} motion-relevant future changes occur, while the global branch provides holistic transition-aware calibration. Consequently, the feedback from imagined futures becomes more spatially selective and informative than using a pooled inverse-dynamics representation alone.

To prevent iterative drift, subsequent closed-loop iterations re-anchor the refined query to the original anchor-conditioned query:
\begin{equation}
e_{k,\mathrm{in}}^{(\ell+1)}
=
\mathrm{MLN}_{\mathrm{cl}}\!\left(
e_{k,\mathrm{init}},
e_{k,\mathrm{ref}}^{(\ell)}
\right),
\qquad
\ell=0,\ldots,L-2.
\end{equation}

The updated query $e_{k,\mathrm{in}}^{(\ell+1)}$ initializes the next
candidate-conditioned rollout, while the anchor $\tau_k$ remains fixed.
Trajectory offsets are decoded only after the final refinement, where
$e_{k,\mathrm{ref}}^{(L-1)}$ is fed into a transformer decoder and an MLP head.
The resulting candidates are ranked by a reward model~\citep{li2025end} to select the final trajectory.

\subsection{Training loss}

Our network is trained with trajectory offset regression, imitation- and simulator-reward supervision, and current/future BEV semantic supervision. We additionally supervise motion correction in the final global IDM feature:
\begin{equation}
\mathcal{L}_{\mathrm{corr}}
=
\ell_{\mathrm{SL1}}\!\left(
h_{\mathrm{mot}}(g_{k^\star}^{(L-1)}),
\operatorname{sg}\!\left[
E_{\mathrm{mot}}(\tau^{\mathrm{GT}})
-
E_{\mathrm{mot}}(\tau_{k^\star})
\right]
\right),
\end{equation}
where $k^\star$ is the closest anchor under winner-take-all assignment, $E_{\mathrm{mot}}$ concatenates fixed normalized speed, acceleration, and heading-rate sequences, and $\operatorname{sg}$ denotes stop-gradient. The masked Smooth L1 loss $\ell_{\mathrm{SL1}}$ averages errors within each motion component and then across components, excluding unreliable low-speed heading rates. The head $h_{\mathrm{mot}}$ is used only during training.

The overall objective is
\begin{equation}
\mathcal{L} = \lambda_{\mathrm{off}}\mathcal{L}_{\mathrm{off}} + \lambda_{\mathrm{off\text{-}im}}\mathcal{L}_{\mathrm{off\text{-}im}} + \lambda_{\mathrm{im}}\mathcal{L}_{\mathrm{im}} + \lambda_{\mathrm{sim}}\mathcal{L}_{\mathrm{sim}} + \lambda_{\mathrm{map}}\mathcal{L}_{\mathrm{map}} + \lambda_{\mathrm{corr}}\mathcal{L}_{\mathrm{corr}},
\end{equation}
where the first five terms supervise offsets, auxiliary and main imitation rewards, simulator metrics, and weighted current/future BEV semantics, respectively. Loss definitions and coefficients are detailed in Appendix~\ref{app:implementation_details}.

\section{Experiments}
\label{others}

\subsection{Benchmark}
We evaluate our method on NAVSIM v1~\citep{dauner2024navsim} and NAVSIM v2~\citep{cao2025pseudo} under the official closed-loop protocols. On NAVSIM v1, we report results on the navtest split using the Predictive Driver Model Score (PDMS), which aggregates no-collision (NC), drivable area compliance (DAC), time-to-collision (TTC), comfort, and ego progress (EP). On NAVSIM v2, we further evaluate on the challenging navhard split under the two-stage pseudo-simulation protocol using the Extended PDMS (EPDMS), which additionally considers driving direction compliance (DDC), traffic light compliance (TLC), lane keeping (LK), history comfort (HC), and extended comfort (EC).

\subsection{Implementation details} 

We use a planning horizon of 4\,s with a sampling interval of 0.5\,s, resulting in 8 future waypoints. Our model adopts a ResNet34-based TransFuser backbone for multi-modal fusion, where the stitched front-view image input is resized to \(256\times1024\). The latent feature dimension is set to 256. In the current implementation, the BEV representation is flattened into 64 latent queries, with \(H_{\mathrm{BEV}}\times W_{\mathrm{BEV}}=8\times8\), and trajectory planning is performed over 256 trajectory anchors. The resulting model contains 69.36M parameters and achieves 17.65 FPS on a single NVIDIA RTX 3090 GPU with batch size 1, indicating that IDOL maintains practical inference efficiency while introducing inverse-dynamics-guided future reasoning. 

Training is conducted on 4 NVIDIA GeForce RTX 3090 GPUs with a batch size of 4 per GPU for approximately 24 hours. We use AdamW with an initial learning rate of $2\times10^{-4}$ and weight decay of $1\times10^{-4}$.

\begin{table}[t] 
\caption{Performance comparison on the NAVSIM v1 navtest split with closed-loop metrics. ``C'' and ``L'' denote camera and LiDAR inputs, respectively. For fair comparison, all listed results are reported under the ResNet-34 image-backbone setting. Bold and underlined values indicate the best and second-best results, respectively. ResWorld$^\ast$ denotes the historical-frame variant that uses historical frames to obtain temporal residuals, and DiffE2E$^\dagger$ denotes the variant with a ResNet-34 image encoder.}
\label{tab:navsim_compare}
\centering
\small
\setlength{\tabcolsep}{3pt}
\renewcommand{\arraystretch}{1.05}
\begin{adjustbox}{max width=\linewidth}
\begin{tabular}{l c c c c c c c c}
\toprule
\textbf{Method} & \textbf{Img. Backbone} & \textbf{Input} & \textbf{NC $\uparrow$} & \textbf{DAC $\uparrow$} & \textbf{TTC $\uparrow$} & \textbf{Comf. $\uparrow$} & \textbf{EP $\uparrow$} & \textbf{PDMS $\uparrow$} \\
\midrule
VADv2~\citep{chen2024vadv2} & ResNet-34 & C \& L & 97.2 & 89.1 & 91.6 & \textbf{100} & 76.0 & \cellcolor{gray!20}80.9 \\
UniAD~\citep{hu2023planning} & ResNet-34 & Camera & 97.8 & 91.9 & 92.9 & \textbf{100} & 78.8 & \cellcolor{gray!20}83.4 \\
PARA-Drive~\citep{weng2024drive} & ResNet-34 & Camera & 97.9 & 92.4 & 93.0 & 99.8 & 79.3 & \cellcolor{gray!20}84.0 \\
TransFuser~\citep{chitta2022transfuser} & ResNet-34 & C \& L & 97.7 & 92.8 & 92.8 & \textbf{100} & 79.2 & \cellcolor{gray!20}84.0 \\
LAW~\citep{li2024enhancing} & ResNet-34 & Camera & 96.4 & 95.4 & 88.7 & \underline{99.9} & 81.7 & \cellcolor{gray!20}84.6 \\
DiffusionDrive~\citep{liao2025diffusiondrive} & ResNet-34 & C \& L & 98.2 & 96.2 & 94.7 & \textbf{100} & 82.2 & \cellcolor{gray!20}88.1 \\
WoTE~\citep{li2025end} & ResNet-34 & C \& L & 98.5 & 96.8 & 94.9 & \underline{99.9} & 81.9 & \cellcolor{gray!20}88.3 \\
SeerDrive~\citep{zhang2025future} & ResNet-34 & C \& L & 98.4 & \underline{97.0} & 94.9 & \underline{99.9} & 83.2 & \cellcolor{gray!20}88.9 \\
ResWorld$^\ast$~\citep{zhang2026resworld} & ResNet-34 & C \& L & \underline{98.9} & 96.5 & \underline{95.6} & \textbf{100} & 83.1 & \cellcolor{gray!20}89.0 \\
MeanFuser~\citep{wang2026meanfuser} & ResNet-34 & Camera & 98.6 & \underline{97.0} & 95.0 & \textbf{100} & 82.8 & \cellcolor{gray!20}89.0 \\
DiffE2E$^\dagger$~\citep{zhao2025diffe2e} & ResNet-34 & C \& L & \textbf{99.2} & 96.8 & \textbf{96.7} & \textbf{100} & \underline{83.6} & \cellcolor{gray!20}\underline{89.8} \\
\midrule
IDOL & ResNet-34 & C \& L & 98.7 & \textbf{97.7} & 95.5 & \textbf{100} & \textbf{84.0} & \cellcolor{gray!20}\textbf{90.0} \\
\bottomrule
\end{tabular}
\end{adjustbox}
\end{table}

\subsection{Comparison with SOTA}

Table~\ref{tab:navsim_compare} compares IDOL with recent end-to-end planners on the NAVSIM navtest split using a ResNet-34~\citep{he2016deep} image backbone. IDOL achieves the best PDMS of 90.0. Table~\ref{tab:navhard_twostage_comparison} further evaluates the challenging NAVSIM v2 navhard split, where IDOL attains the highest final EPDMS of 38.0, outperforming the strongest baseline  WoTE by 10.1 points.

Beyond the main comparisons, Table~\ref{tab:navsimv2_navtest_epdms} reports the extended NAVSIM v2 navtest benchmark. IDOL achieves 89.6 EPDMS, ranking first among learned methods and remaining competitive with the human agent. Table~\ref{tab:navhard_stage1_epdms} further evaluates the more challenging navhard stage-1-only setting, where IDOL achieves the best performance with 76.2 EPDMS, surpassing Hydra-MDP$^\dagger$ by 3.1 points and WoTE by 9.5 points. Across NAVSIM v1 navtest, NAVSIM v2 navtest, navhard stage-1, and the two-stage navhard evaluation, IDOL consistently achieves state-of-the-art performance among comparable learned planners, demonstrating the robustness and stability of inverse-dynamics-guided future reasoning under both standard and challenging long-tail driving scenarios.

\begin{table*}[t]
\caption{Performance comparison on the NAVSIM v2 navhard split. PDM-Closed is listed separately as a privileged planner using ground-truth perception.
}
\label{tab:navhard_twostage_comparison}
\centering
\small
\setlength{\tabcolsep}{3.2pt}
\renewcommand{\arraystretch}{1.08}
\resizebox{\textwidth}{!}{
\begin{tabular}{l c c c c c c c c c c c c}
\toprule
\textbf{Method} & \textbf{Backbone} & \textbf{Stage} &
\textbf{NC $\uparrow$} & \textbf{DAC $\uparrow$} & \textbf{DDC $\uparrow$} & \textbf{TLC $\uparrow$} &
\textbf{EP $\uparrow$} & \textbf{TTC $\uparrow$} & \textbf{LK $\uparrow$} & \textbf{HC $\uparrow$} & \textbf{EC $\uparrow$} &
\textbf{EPDMS $\uparrow$} \\
\midrule
\multirow{2}{*}{PDM-Closed~\citep{dauner2023parting}}
& \multirow{2}{*}{-}
& Stage 1 & 94.4 & 98.8 & 100.0 & 99.5 & 100.0 & 93.5 & 99.3 & 87.7 & 36.0 & \cellcolor{gray!20} \\
& & Stage 2 & 88.1 & 90.6 & 96.3 & 98.5 & 100.0 & 83.1 & 73.7 & 91.5 & 25.4 & \cellcolor{gray!20}\multirow{-2}{*}{51.3} \\
\midrule
\multirow{2}{*}{LTF~\citep{prakash2021multi}}
& \multirow{2}{*}{ResNet34}
& Stage 1 & 96.2 & 79.5 & \textbf{99.1} & 99.5 & \textbf{84.1} & 95.1 & 94.2 & 97.5 & \textbf{79.1} & \cellcolor{gray!20} \\
& & Stage 2 & 77.7 & 70.2 & 84.2 & 98.0 & 85.1 & 75.6 & 45.4 & 95.7 & \textbf{75.9} & \cellcolor{gray!20}\multirow{-2}{*}{23.1} \\
\midrule
\multirow{2}{*}{GTRS-DP~\citep{li2025generalized}}
& \multirow{2}{*}{ResNet34}
& Stage 1 & 94.7 & 78.8 & 96.1 & 99.5 & 83.0 & 94.4 & 92.0 & 97.5 & 72.8 & \cellcolor{gray!20} \\
& & Stage 2 & 80.3 & 74.4 & 84.9 & 98.0 & 81.9 & 78.8 & 45.4 & \textbf{96.7} & 70.1 & \cellcolor{gray!20}\multirow{-2}{*}{23.8} \\
\midrule
\multirow{2}{*}{DiffusionDrive~\citep{liao2025diffusiondrive}}
& \multirow{2}{*}{ResNet34}
& Stage 1 & 96.0 & 79.7 & 97.4 & 99.5 & 81.3 & 93.1 & 90.8 & 96.8 & 73.8 & \cellcolor{gray!20} \\
& & Stage 2 & 82.1 & 72.2 & 88.5 & 98.7 & 85.1 & 78.8 & 49.2 & 89.3 & 71.2 & \cellcolor{gray!20}\multirow{-2}{*}{24.2} \\
\midrule
\multirow{2}{*}{GuideFlow~\citep{liu2025guideflow}}
& \multirow{2}{*}{ResNet34}
& Stage 1 & 96.6 & 80.5 & 96.3 & 99.3 & 82.3 & 94.9 & 91.5 & \textbf{97.7} & 67.8 & \cellcolor{gray!20} \\
& & Stage 2 & \textbf{87.3} & 76.7 & \textbf{88.8} & \textbf{99.2} & 84.3 & \textbf{85.1} & 49.7 & 93.1 & 44.5 & \cellcolor{gray!20}\multirow{-2}{*}{27.1} \\
\midrule
\multirow{2}{*}{WoTE~\citep{li2025end}}
& \multirow{2}{*}{ResNet34}
& Stage 1 & \textbf{97.4} & 88.2 & 97.7 & 99.3 & 82.7 & 96.4 & 90.8 & 97.3 & 68.0 & \cellcolor{gray!20} \\
& & Stage 2 & 81.2 & 77.7 & 84.8 & 98.1 & \textbf{85.9} & 78.5 & 46.2 & 96.6 & 63.3 & \cellcolor{gray!20}\multirow{-2}{*}{27.9} \\
\midrule
\multirow{2}{*}{IDOL}
& \multirow{2}{*}{ResNet34}
& Stage 1 & 97.2 & \textbf{89.6} & 98.0 & \textbf{99.6} & 82.3 & \textbf{96.9} & \textbf{95.3} & 97.6 & 72.9 & \cellcolor{gray!20} \\
& & Stage 2 & 84.9 & \textbf{81.7} & 87.7 & 98.5 & 84.7 & 81.1 & \textbf{50.0} & 95.4 & 64.0 & \cellcolor{gray!20}\multirow{-2}{*}{\textbf{38.0}} \\
\bottomrule
\end{tabular}
}
\end{table*}

\subsection{Ablation study}

We conduct ablation studies on the NAVSIM navtest split to analyze the contribution
of each design in IDOL. We report PDMS and its five component metrics. Additional ablation results and analyses are provided in Appendix~\ref{sec:appendix_additional_ablation}.

\paragraph{Effect of inverse dynamics and closed-loop refinement.}
Table~\ref{tab:ablation_main} studies the main components of IDOL. Without IDM,
future BEV states lack explicit transition-to-motion feedback, limiting their effect on
trajectory refinement. Introducing IDM improves PDMS from 87.3 to 89.2 by converting
latent future transitions into motion-aware query updates. A single closed-loop refinement
iteration further raises PDMS to 90.0 by feeding the updated query back into future
reasoning. Increasing the number of iterations to two or three reduces performance,
suggesting possible over-correction.

\begin{table}[ht]
\caption{Main ablation of inverse dynamics and closed-loop refinement on the NAVSIM navtest split.
IDM denotes the inverse dynamics model, and CL denotes closed-loop refinement.}
\label{tab:ablation_main}
\centering
\small
\setlength{\tabcolsep}{5.0pt}
\renewcommand{\arraystretch}{1.05}
\begin{tabular}{c c c c c c c c}
\toprule
\textbf{IDM} & \textbf{CL iters.} & \textbf{NC} $\uparrow$ & \textbf{DAC} $\uparrow$ & \textbf{TTC} $\uparrow$ & \textbf{Comf.} $\uparrow$ & \textbf{EP} $\uparrow$ & \textbf{PDMS} $\uparrow$ \\
\midrule
$\times$ & $\times$ & 98.5 & 95.0 & 95.2 & \textbf{100} & 81.5 & \cellcolor{gray!20}87.3 \\
\checkmark & $\times$ & 98.6 & 97.0 & 95.4 & \textbf{100} & 83.3 & \cellcolor{gray!20}89.2 \\
\checkmark & $1$ & 98.7 & \textbf{97.7} & 95.5 & \textbf{100} & \textbf{84.0} & \cellcolor{gray!20}\textbf{90.0} \\
\checkmark & $2$ & 98.6 & 97.3 & 95.2 & \textbf{100} & 83.5 & \cellcolor{gray!20}89.4 \\
\checkmark & $3$ & \textbf{98.8} & 97.2 & \textbf{95.7} & \textbf{100} & 83.1 & \cellcolor{gray!20}89.4 \\
\bottomrule
\end{tabular}
\end{table}

\paragraph{Temporal design of IDM.}
Table~\ref{tab:ablation_temporal} compares different temporal inputs for IDM. The
two-frame design directly decodes adjacent BEV transitions, while the four-frame variant
uses a longer window that may dilute local transition cues for immediate refinement.
The better performance of the two-frame setting shows that adjacent transitions provide
a more compact and effective basis for planning-oriented motion refinement.

\begin{table}[ht]
\caption{Ablation of temporal decoding in the inverse dynamics model. The two-frame
variant decodes adjacent BEV transitions, while the four-frame variant uses a longer
future window.}
\label{tab:ablation_temporal}
\centering
\small
\setlength{\tabcolsep}{5.0pt}
\renewcommand{\arraystretch}{1.05}
\begin{tabular}{l c c c c c c}
\toprule
\textbf{IDM temporal input} & \textbf{NC} $\uparrow$ & \textbf{DAC} $\uparrow$ & \textbf{TTC} $\uparrow$ & \textbf{Comf.} $\uparrow$ & \textbf{EP} $\uparrow$ & \textbf{PDMS} $\uparrow$ \\
\midrule
4 frame & \textbf{98.7} & 97.3 & \textbf{95.6} & \textbf{100} & 83.7 & \cellcolor{gray!20}89.6 \\
2 frame & \textbf{98.7} & \textbf{97.7} & 95.5 & \textbf{100} & \textbf{84.0} & \cellcolor{gray!20}\textbf{90.0} \\
\bottomrule
\end{tabular}
\end{table}

\paragraph{Effect of spatial and global dynamics branches.}
Table~\ref{tab:ablation_branch} analyzes the dual-branch design. The global branch
captures holistic transition patterns but lacks spatial selectivity, while the spatial
branch preserves local transition evidence but lacks global calibration. Combining both
branches provides spatially selective and globally calibrated future-to-planning feedback,
achieving the best overall PDMS.

\begin{table}[ht]
\caption{Ablation of the spatial and global dynamics branches in the inverse-dynamics-guided
refinement network.}
\label{tab:ablation_branch}
\centering
\small
\setlength{\tabcolsep}{5.0pt}
\renewcommand{\arraystretch}{1.05}
\begin{tabular}{l c c c c c c}
\toprule
\textbf{Dynamics Branch} & \textbf{NC} $\uparrow$ & \textbf{DAC} $\uparrow$ & \textbf{TTC} $\uparrow$ & \textbf{Comf.} $\uparrow$ & \textbf{EP} $\uparrow$ & \textbf{PDMS} $\uparrow$ \\
\midrule
w/o spatial branch & \textbf{98.8} & 96.9 & \textbf{95.8} & \textbf{100} & 83.2 & \cellcolor{gray!20}89.3 \\
w/o global branch & 98.7 & 97.2 & 95.3 & \textbf{100} & 83.9 & \cellcolor{gray!20}89.5 \\
dual-branch IDM & 98.7 & \textbf{97.7} & 95.5 & \textbf{100} & \textbf{84.0} & \cellcolor{gray!20}\textbf{90.0} \\
\bottomrule
\end{tabular}
\end{table}

\subsection{Visualization}
\begin{figure}[htbp]
    \centering
    \includegraphics[width=\linewidth]{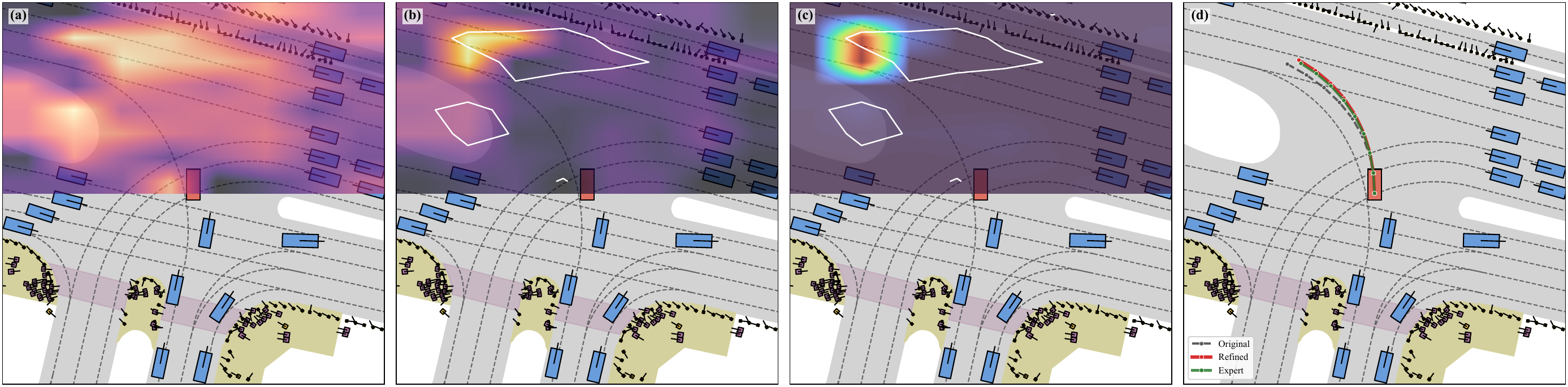}
    \caption{Qualitative visualization of inverse-dynamics-guided refinement on the NAVSIM navtest split. }
    \label{fig:visual}
\end{figure}

To better understand inverse-dynamics-guided refinement, we visualize a representative NAVSIM navtest case in Figure~\ref{fig:visual}. Panels (a)--(c) show the latent BEV change, the corresponding IDM spatial response, and the refinement-query attention, respectively. The white contour marks high-change regions extracted from Panel (a). The visualization shows that IDM indeed concentrates on the motion-informative parts within the latent BEV change regions, rather than unrelated background variations, and the query attends to these motion-aware cues during refinement.  Panel (d) compares the original, refined, and expert trajectories, where the refined trajectory better aligns with the expert, suggesting that IDM converts latent future transitions into actionable planning updates. More qualitative results are provided in Appendix~\ref{sec:Additional_qualitative}.

\section{Conclusion}

In this work, we proposed IDOL, an inverse-dynamics-guided future prediction framework for end-to-end autonomous driving. By decoding adjacent latent BEV futures, IDOL converts predicted scene evolution into transition-aware motion cues and refines trajectories through a lightweight closed-loop process. This explicitly bridges world modeling and executable motion generation, making future prediction more actionable for planning. Experiments on NAVSIM v1 and NAVSIM v2 show that IDOL achieves state-of-the-art performance on navtest, navhard stage-1, and two-stage navhard benchmarks, demonstrating the effectiveness and generality of inverse-dynamics-guided future reasoning. This framework provides a general paradigm for integrating future state transitions with motion generation in future autonomous driving research.

\bibliography{iclr2027_conference}
\bibliographystyle{iclr2027_conference}

\clearpage
\appendix

\section{Notation}
\label{sec:app_notation}

Table~\ref{tab:app_notation} summarizes the main notation used in IDOL. We use $k$ to index trajectory candidates, $s$ to index trajectory poses, $u$ to index latent rollout steps, and $\ell$ to index query-refinement iterations. The final refinement iteration is $\ell=L-1$.

\begin{table}[htbp]
\centering
\small
\setlength{\tabcolsep}{5pt}
\renewcommand{\arraystretch}{1.12}
\caption{Main notation used in IDOL.}
\label{tab:app_notation}
\begin{tabular}{l p{0.68\linewidth}}
\toprule
\textbf{Symbol} & \textbf{Description} \\
\midrule
$K,\ S$
& Number of trajectory anchors and future poses per trajectory. \\
$U,\ L$
& Number of latent rollout steps and query-refinement iterations. \\
$B$
& Current latent BEV representation. \\
$\mathcal{A}=\{\tau_k\}_{k=1}^{K}$
& Trajectory-anchor vocabulary, with $\tau_k\in\mathbb{R}^{S\times 3}$ containing planar positions and headings. \\
$\tau^{\mathrm{GT}}$
& Expert future trajectory in the current ego frame. \\
$a_k,\ z$
& Encoded anchor feature and ego-state feature. \\
$e_{k,\mathrm{init}}$
& Initial planning query conditioned on the ego state and anchor $k$. \\
$e_{k,\mathrm{in}}^{(\ell)},\ e_{k,\mathrm{ref}}^{(\ell)}$
& Planning query before and after refinement at iteration $\ell$. \\
$q_{k,\ell}^{(u)}$
& Ego query at step $u$ of the candidate-conditioned latent rollout. \\
$B_{k,\ell,\mathrm{cond}}^{(u)}$
& Latent BEV state conditioned through ego-query injection. \\
$S_k^{(\ell)},\ g_k^{(\ell)}$
& Spatial and global IDM features aggregated over rollout transitions. \\
$k^\star$
& Index of the anchor closest to $\tau^{\mathrm{GT}}$ under the winner-take-all assignment. \\
$E_{\mathrm{mot}}(\tau)$
& Fixed, normalized motion representation of trajectory $\tau$. \\
$h_{\mathrm{mot}}$
& Training-only head predicting a motion-feature residual from the final global IDM feature. \\
$\operatorname{sg}[\cdot]$
& Stop-gradient operator applied to the supervision target. \\
\bottomrule
\end{tabular}
\end{table}

\section{Implementation details}
\label{app:implementation_details}

\paragraph{Hyperparameters.}
Table~\ref{tab:app_important_hyperparams} summarizes the key hyperparameters used in IDOL. The trajectory-anchor vocabulary is constructed offline and kept fixed during training and evaluation. For ablation studies, all settings are unchanged unless the corresponding component is explicitly modified.

\begin{table}[htbp]
\centering
\small
\setlength{\tabcolsep}{5pt}
\renewcommand{\arraystretch}{1.12}
\caption{Hyperparameters of IDOL.}
\label{tab:app_important_hyperparams}
\begin{tabular}{l c}
\toprule
\textbf{Parameter} & \textbf{Value} \\
\midrule
Epochs & 30 \\
Image input size & $256 \times 1024$ \\
BEV latent resolution & $8 \times 8$ \\
Number of BEV queries & 64 \\
Latent feature dimension & 256 \\
Planning horizon & 4 s \\
Waypoint interval & 0.5 s \\
Number of future waypoints & 8 \\
Trajectory-anchor vocabulary size & 256 \\
Closed-loop refinement iterations & 1 \\
Learning rate & $2 \times 10^{-4}$ \\ 
Weight decay & $1 \times 10^{-4}$ \\
\bottomrule
\end{tabular}
\end{table}

\paragraph{Inverse dynamics model.}
IDOL adopts a lightweight MLP-based inverse dynamics model to convert adjacent imagined BEV states into motion-aware transition cues.
Given two consecutive latent BEV states, the model first forms a transition representation by concatenating their query features along the channel dimension.
A three-stage MLP transition encoder is then applied: an initial fusion stage compresses the paired-state representation into the latent feature space, followed by residual MLP transformations that refine query-wise transition patterns while preserving the original spatial layout.
The resulting query-wise representation can be viewed as a \emph{spatial dynamics map}, which captures local motion-relevant changes across BEV queries.
IDOL further aggregates this spatial dynamics map into a compact \emph{global dynamics feature}, which serves as a trajectory-level motion condition for modulating the anchor-conditioned ego planning feature through an MLN layer before trajectory offset prediction.
This design keeps the inverse dynamics branch lightweight while allowing the planner to exploit both local transition patterns and global motion context from adjacent imagined futures.

\paragraph{Loss configuration.}
IDOL is trained with trajectory offset regression, reward supervision, BEV semantic supervision, and IDM motion-feature correction. Table~\ref{tab:app_loss_terms} lists the loss coefficients, which remain fixed throughout training.

\begin{table}[htbp]
\centering
\small
\setlength{\tabcolsep}{4pt}
\renewcommand{\arraystretch}{1.12}
\caption{Training losses and their coefficients.}
\label{tab:app_loss_terms}
\begin{tabular}{l l c}
\toprule
\textbf{Loss term} & \textbf{Role} & \textbf{Coefficient} \\
\midrule
$\mathcal{L}_{\mathrm{off}}$
& Trajectory offset regression & $1.0$ \\
$\mathcal{L}_{\mathrm{off\text{-}im}}$
& Auxiliary imitation-reward supervision & $0.1$ \\
$\mathcal{L}_{\mathrm{im}}$
& Main imitation-reward supervision & $1.0$ \\
$\mathcal{L}_{\mathrm{sim}}$
& Simulator-metric reward supervision & $1.0$ \\
$\mathcal{L}_{\mathrm{map}}^{\mathrm{cur}}$
& Current BEV semantic supervision & $10.0$ \\
$\mathcal{L}_{\mathrm{map}}^{\mathrm{fut}}$
& Future BEV semantic supervision & $0.1$ \\
$\mathcal{L}_{\mathrm{corr}}$
& IDM motion-feature correction & $0.2$ \\
\bottomrule
\end{tabular}
\end{table}

The offset loss $\mathcal{L}_{\mathrm{off}}$ applies L1 regression to the trajectory correction of the anchor closest to the expert trajectory under winner-take-all assignment.
The imitation losses $\mathcal{L}_{\mathrm{off\text{-}im}}$ and $\mathcal{L}_{\mathrm{im}}$ supervise the auxiliary and main reward heads using a soft target distribution derived from anchor-to-expert distances.
The simulator loss $\mathcal{L}_{\mathrm{sim}}$ sums binary cross-entropy losses over five cached planning metrics.
The current and future BEV losses supervise semantic scene representations using focal loss with $\alpha=0.5$ and $\gamma=2$.

Finally, downstream planning losses supervise the IDM indirectly, leaving the motion information encoded in its features underconstrained. We introduce $\mathcal{L}_{\mathrm{corr}}$ to explicitly encourage the final global IDM feature to encode the motion adjustment from a candidate anchor toward the expert trajectory. For a trajectory $\tau$, we define the fixed motion representation
\begin{equation}
E_{\mathrm{mot}}(\tau)
=
\left[
v_{1:S}/\sigma_v,\;
a_{1:S-1}/\sigma_a,\;
\omega_{1:S}/\sigma_\omega
\right],
\end{equation}
where $v_s$ is the signed displacement projected onto the midpoint heading and divided by $\Delta t$, $a_s=(v_{s+1}-v_s)/\Delta t$, and $\omega_s$ is the wrapped heading increment divided by $\Delta t$. We prepend the current ego pose and use $\Delta t=0.5\,\mathrm{s}$, $S=8$, and normalization scales $(\sigma_v,\sigma_a,\sigma_\omega)=(10\,\mathrm{m/s},3\,\mathrm{m/s^2},0.5\,\mathrm{rad/s})$.

Let $k^\star$ denote the anchor closest to the expert trajectory under the same winner-take-all assignment as offset regression. The correction loss is
\begin{equation}
\mathcal{L}_{\mathrm{corr}}
=
\ell_{\mathrm{SL1}}\!\left(
h_{\mathrm{mot}}(g_{k^\star}^{(L-1)}),
\operatorname{sg}\!\left[
E_{\mathrm{mot}}(\tau^{\mathrm{GT}})
-
E_{\mathrm{mot}}(\tau_{k^\star})
\right]
\right),
\end{equation}
where $h_{\mathrm{mot}}$ is an auxiliary MLP head and $\operatorname{sg}$ denotes stop-gradient. The operator $\ell_{\mathrm{SL1}}$ computes masked Smooth L1 loss with unit transition threshold, averaging valid entries within each motion component and then averaging the three components. Invalid targets are excluded, and heading-rate supervision is masked when either the expert or winning-anchor interval speed has magnitude below $0.5\,\mathrm{m/s}$. Only the winning candidate is supervised. This fixed kinematic target provides interpretable supervision for the IDM representation used in query refinement. We use $\lambda_{\mathrm{corr}}=0.2$, and the auxiliary head is discarded at inference.

\newcommand{\best}[1]{\textbf{#1}}
\newcommand{\second}[1]{\underline{#1}}
\newcommand{\epdmscell}[1]{\cellcolor{gray!20}{#1}}

\begin{table*}[tbp] 
\caption{
Performance comparison on the NAVSIM v2 navtest split using extended closed-loop metrics.
The best and second-best results among learned methods are highlighted in bold and underlined, respectively.
Superscripts indicate methods using additional training signals or settings beyond standard imitation learning:
$^\ast$ denotes VLA/VLM-based methods,
$^\dagger$ denotes reinforcement learning, reward-guided training, or reinforcement fine-tuning,
and $^\ddagger$ denotes extra synthetic, pseudo-expert, or QA-style data.
}
\label{tab:navsimv2_navtest_epdms}
\centering
\small
\setlength{\tabcolsep}{3.0pt}
\renewcommand{\arraystretch}{1.06}
\resizebox{\textwidth}{!}{
\begin{tabular}{l c c c c c c c c c c}
\toprule
\textbf{Method} &
\textbf{NC$\uparrow$} &
\textbf{DAC$\uparrow$} &
\textbf{DDC$\uparrow$} &
\textbf{TLC$\uparrow$} &
\textbf{EP$\uparrow$} &
\textbf{TTC$\uparrow$} &
\textbf{LK$\uparrow$} &
\textbf{HC$\uparrow$} &
\textbf{EC$\uparrow$} &
\textbf{EPDMS$\uparrow$} \\
\midrule
Human Agent
& 100.0 & 100.0 & 99.8 & 100.0 & 87.4 & 100.0 & 100.0 & 98.1 & 90.1 & \epdmscell{90.3} \\
\midrule
Ego Status MLP~\citep{li2024ego}
& 93.1 & 77.9 & 92.7 & 99.6 & 86.0 & 91.5 & 89.4 & \best{98.3} & 85.4 & \epdmscell{64.0} \\
TransFuser~\citep{chitta2022transfuser}
& 96.9 & 89.9 & 97.8 & 99.7 & 87.1 & 95.4 & 92.7 & \best{98.3} & 87.2 & \epdmscell{76.7} \\
Hydra-MDP++~\citep{li2025hydramdpp}
& 97.2 & 97.5 & 99.4 & 99.6 & 83.1 & 96.5 & 94.4 & \second{98.2} & 70.9 & \epdmscell{81.4} \\
DriveSuprim~\citep{yao2026drivesuprim}
& 97.5 & 96.5 & 99.4 & 99.6 & 88.4 & 96.6 & 95.5 & \best{98.3} & 77.0 & \epdmscell{83.1} \\
ReCogDrive$^{\ast,\dagger}$~\citep{li2025recogdrive}
& 98.3 & 95.2 & 99.5 & \second{99.8} & 87.1 & 97.5 & 96.6 & \best{98.3} & 86.5 & \epdmscell{83.6} \\
DiffusionDrive~\citep{liao2025diffusiondrive}
& 98.2 & 95.9 & 99.4 & \second{99.8} & 87.5 & 97.3 & 96.8 & \best{98.3} & 87.7 & \epdmscell{84.5} \\
GTRS-Dense + SimScale$^{\ddagger,\dagger}$~\citep{tian2025simscale}
& 98.4 & 98.8 & 99.4 & \best{99.9} & 87.9 & 98.1 & 96.4 & 97.6 & 58.8 & \epdmscell{84.6} \\
World4Drive~\citep{zheng2025world4drive}
& 97.8 & 96.3 & 99.4 & \second{99.8} & 88.3 & 97.1 & \best{97.7} & 98.0 & 53.9 & \epdmscell{84.8} \\
Epona~\citep{zhang2025epona}
& 97.1 & 95.7 & 99.3 & 99.7 & \second{88.6} & 96.3 & 97.0 & 98.0 & 67.8 & \epdmscell{85.1} \\
DiffusionDriveV2$^\dagger$~\citep{zou2025diffusiondrivev2}
& 97.7 & 96.6 & 99.2 & \second{99.8} & \best{88.9} & 97.2 & 96.0 & 97.8 & \best{91.0} & \epdmscell{85.5} \\
VADv2~\citep{chen2024vadv2}
& 98.0 & 98.3 & 99.4 & \second{99.8} & 87.1 & 96.8 & 95.2 & \best{98.3} & 88.1 & \epdmscell{85.8} \\
DriveVLA-W0$^\ast$~\citep{li2025drivevla}
& 98.5 & \best{99.1} & 98.0 & 99.7 & 86.4 & 98.1 & 93.2 & 97.9 & 58.9 & \epdmscell{86.1} \\
SGDrive$^{\ast,\ddagger}$~\citep{li2026sgdrive}
& 98.6 & 94.3 & 99.5 & \best{99.9} & 86.0 & 97.9 & 96.1 & \best{98.3} & 85.9 & \epdmscell{86.2} \\
DiffRefiner~\citep{yin2026diffrefiner}
& 98.5 & 97.4 & \second{99.6} & \second{99.8} & 87.6 & 97.7 & \best{97.7} & \best{98.3} & 86.2 & \epdmscell{86.2} \\
WorldRFT$^\dagger$~\citep{yang2026worldrft}
& 97.8 & 96.5 & 99.5 & \second{99.8} & 88.5 & 97.0 & 97.4 & 98.1 & 69.1 & \epdmscell{86.7} \\
SafeDrive~\citep{kim2026safedrive}
& \best{99.5} & \second{99.0} & \best{99.7} & \best{99.9} & \second{88.6} & \best{98.9} & \second{97.5} & \second{98.2} & 81.9 & \epdmscell{87.5} \\
MeanFuser~\citep{wang2026meanfuser}
& 98.3 & 97.2 & \second{99.6} & \second{99.8} & 87.6 & 97.4 & 97.3 & \best{98.3} & \second{88.2} & \epdmscell{\second{89.5}} \\
\midrule
IDOL
& \second{98.8} & 97.6 & 99.5 & \second{99.8} & 87.1 & \second{98.3} & 96.3 & \best{98.3} & 85.5 & \epdmscell{\best{89.6}} \\
\bottomrule
\end{tabular}
}
\end{table*}

\section{Benchmark and metrics details}
\label{app:datasets_metrics}

\paragraph{Benchmark.}
All models are trained on the NAVSIM training split. We evaluate IDOL on the NAVSIM benchmark. The \textit{navtest} split is the standard test split used to evaluate closed-loop planning performance under common urban driving scenarios. The \textit{navhard} split is a more challenging subset designed to stress-test planners under difficult cases such as complex interactions, stricter route following, and long-tail driving conditions. For NAVSIM v2, navhard is evaluated under the pseudo-simulation protocol, including the stage-1-only setting and the two-stage setting. The former measures the quality of the initially generated trajectory, while the latter further evaluates planning after pseudo-simulation rollout and therefore reflects closed-loop robustness more directly. Since point-cloud inputs are unavailable in stage 2, we separately train a camera-based IDOL variant using the Latent TransFuser (LTF) formulation. Specifically, the LiDAR input is replaced by a learnable latent tensor initialized from an LTF checkpoint and jointly optimized with the IDOL network on the NAVSIM training split, while retaining the fusion architecture and downstream planning modules. The resulting IDOL-LTF checkpoint is used throughout the two-stage evaluation; the model does not switch modalities between stages or rely on dynamic modality-dropout training.

\paragraph{PDMS.}
For NAVSIM v1, we report the Predictive Driver Model Score (PDMS), which evaluates trajectory quality after non-reactive simulation. PDMS uses no-collision (NC) and drivable area compliance (DAC) as multiplicative safety constraints, and combines ego progress (EP), time-to-collision within bound (TTC), and comfort (C) as weighted quality terms:
\begin{equation}
\mathrm{PDMS}
=
\mathrm{NC}
\cdot
\mathrm{DAC}
\cdot
\frac{
5\mathrm{EP}
+
5\mathrm{TTC}
+
2\mathrm{C}
}{12}.
\end{equation}
All component metrics are normalized to $[0,1]$, and the final reported PDMS is averaged over all evaluated scenarios and expressed as a percentage.

\paragraph{EPDMS.}
For NAVSIM v2, we report the Extended Predictive Driver Model Score (EPDMS). Compared with PDMS, EPDMS adds driving direction compliance (DDC) and traffic light compliance (TLC) as multiplicative rule-compliance constraints, and further introduces lane keeping (LK), history comfort (HC), and extended comfort (EC) into the weighted quality term. Following the NAVSIM v2 metric definition, we write the filtered value of each metric as $\tilde{m}$, where human-trajectory violations are filtered to avoid false-positive penalties. EPDMS is computed as
\begin{equation}
\mathrm{EPDMS}
=
\tilde{\mathrm{NC}}
\cdot
\tilde{\mathrm{DAC}}
\cdot
\tilde{\mathrm{DDC}}
\cdot
\tilde{\mathrm{TLC}}
\cdot
\frac{
5\tilde{\mathrm{EP}}
+
5\tilde{\mathrm{TTC}}
+
2\tilde{\mathrm{LK}}
+
2\tilde{\mathrm{HC}}
+
2\tilde{\mathrm{EC}}
}{16}.
\end{equation}
EPDMS therefore extends PDMS from basic safety and progress evaluation to stricter rule compliance, lane keeping, and pseudo-closed-loop consistency, making it more suitable for NAVSIM v2 and the challenging navhard split.

\section{Additional evaluation on NAVSIM v2 and Bench2Drive}
\label{sec:appendix_navsim_v2}

\paragraph{Evaluation on NAVSIM v2.}
We report extended NAVSIM v2 results on the standard navtest split and the more challenging navhard stage-1 setting. Additional baseline results are collected from publicly available papers~\citep{jiang2025irl,sun2025minddrive,sun2026diffsemanticfusion,liu2025guideflow}. As shown in Table~\ref{tab:navsimv2_navtest_epdms}, IDOL achieves 89.6 EPDMS on navtest, ranking first among learned methods and remaining competitive with the human agent, with consistently strong safety, rule-compliance, and comfort-related metrics.

Under the navhard stage-1 protocol in Table~\ref{tab:navhard_stage1_epdms}, IDOL achieves 76.2 EPDMS, outperforming prior learned planners including Hydra-MDP and WoTE. These results support the effectiveness of inverse-dynamics-guided future reasoning across both standard and challenging long-tail NAVSIM v2 settings.

\paragraph{Interactive closed-loop evaluation on Bench2Drive.}
To evaluate performance beyond NAVSIM, we additionally test IDOL on Bench2Drive, a CARLA-based benchmark comprising 220 routes and 44 safety-critical interactive scenarios. Following the official evaluation protocol, we report Driving Score (DS) and Success Rate (SR) to assess driving performance and task completion under continuous interaction with the environment.

As shown in Table~\ref{tab:bench2drive_results}, IDOL uses a ResNet-34 image backbone and achieves a DS of 63.22 and an SR of 34.09\%, outperforming the listed planners, including the future-BEV-based methods SeerDrive and WoTE. Compared with WoTE, IDOL improves DS by 1.51 points and SR by 2.73 percentage points. These results provide further evidence that inverse-dynamics-guided future reasoning supports effective planning and task completion under interactive closed-loop evaluation beyond NAVSIM.

\begin{table*}[htbp]
\caption{
Performance comparison on the NAVSIM v2 navhard split under the stage-1-only protocol.
Bold and underlined values indicate the best and second-best results, respectively, for each metric.
$^\dagger$ indicates results reported with the V2-99 backbone.
}
\label{tab:navhard_stage1_epdms}
\centering
\scriptsize
\setlength{\tabcolsep}{3.2pt}
\renewcommand{\arraystretch}{1.05}
\begin{tabular}{lcccccccccc}
\toprule
\textbf{Method} &
\textbf{NC$\uparrow$} &
\textbf{DAC$\uparrow$} &
\textbf{DDC$\uparrow$} &
\textbf{TLC$\uparrow$} &
\textbf{EP$\uparrow$} &
\textbf{TTC$\uparrow$} &
\textbf{LK$\uparrow$} &
\textbf{HC$\uparrow$} &
\textbf{EC$\uparrow$} &
\textbf{EPDMS$\uparrow$} \\
\midrule
TransFuser~\citep{chitta2022transfuser}
& 96.3
& 74.6
& 98.4
& 99.3
& 82.9
& 93.7
& 92.7
& 97.5
& \underline{78.2}
& \cellcolor{gray!20} 60.5 \\

LTF~\citep{chitta2022transfuser}
& 96.2
& 79.5
& \underline{99.1}
& 99.5
& \underline{84.1}
& 95.1
& 94.2
& 97.5
& \textbf{79.1}
& \cellcolor{gray!20} 62.3 \\

DiffusionDrive~\citep{liao2025diffusiondrive}
& 95.9
& 84.0
& 98.6
& \textbf{99.8}
& \textbf{84.4}
& 96.0
& \underline{95.1}
& \underline{97.6}
& 71.1
& \cellcolor{gray!20} 63.2 \\

WoTE~\citep{li2025end}
& \underline{97.4}
& 88.2
& 97.8
& 99.3
& 82.7
& 96.4
& 90.9
& 97.3
& 68.0
& \cellcolor{gray!20} 66.7 \\

Hydra-MDP$^\dagger$~\citep{li2024hydra}
& \textbf{97.6}
& \textbf{96.4}
& \textbf{99.2}
& 99.3
& 80.2
& \textbf{96.9}
& 94.9
& \textbf{97.8}
& 58.7
& \cellcolor{gray!20} \underline{73.1} \\

\midrule
IDOL
& 97.2
& \underline{89.6}
& 98.0
& \underline{99.6}
& 82.3
& \textbf{96.9}
& \textbf{95.3}
& \underline{97.6}
& 72.9
& \cellcolor{gray!20} \textbf{76.2} \\
\bottomrule
\end{tabular}
\end{table*}

\begin{table}[htbp]
\caption{
Interactive closed-loop evaluation on Bench2Drive.
Higher Driving Score (DS) and Success Rate (SR) indicate better performance.
Bold values denote the best results among the listed methods.
}
\label{tab:bench2drive_results}
\centering
\small
\setlength{\tabcolsep}{10pt}
\renewcommand{\arraystretch}{1.05}
\begin{tabular}{l c c}
\toprule
\textbf{Method} & \textbf{Driving Score (DS)} $\uparrow$ & \textbf{Success Rate (SR)} $\uparrow$ \\
\midrule
VAD~\citep{jiang2023vad} & 42.35 & \cellcolor{gray!20}15.00\% \\
UniAD-Base~\citep{hu2023planning} & 45.81 & \cellcolor{gray!20}16.36\% \\
SeerDrive~\citep{zhang2025future} & 58.32 & \cellcolor{gray!20}30.17\% \\
WoTE~\citep{li2025end} & 61.71 & \cellcolor{gray!20}31.36\% \\
IDOL & \textbf{63.22} & \cellcolor{gray!20}\textbf{34.09\%} \\
\bottomrule
\end{tabular}
\end{table}

\section{Additional ablation studies}
\label{sec:appendix_additional_ablation}

We further examine whether learned inverse dynamics provides better planning guidance than simpler future-state representations, whether its gains arise from additional model capacity, how global inverse dynamics features should be fused with the refined planning query, and how performance varies with the trajectory-anchor vocabulary size.

\paragraph{Ablation on transition-aware future modeling.}
We compare three representations for planning refinement: imagined future BEV states without explicit transition modeling, direct latent differences between adjacent future states, and IDM features learned from ordered pairs of adjacent states.
To control for model capacity, we additionally apply parameter-matched MLPs to the future-state-only and latent-difference representations.
Both augmented variants have exactly the same total parameter count as IDOL.
Across these comparisons, the world model, spatial cross-attention, global modulation, iterative refinement, trajectory anchors, losses, and training settings remain unchanged.

As shown in Table~\ref{tab:app_transition_ablation}, latent differences improve PDMS from 88.6 to 89.0 over future states alone, supporting the value of transition information.
Adding parameter-matched MLPs yields only modest gains of 0.3 and 0.1 points, respectively.
IDM achieves 90.0 PDMS, exceeding the two parameter-matched variants by 1.1 and 0.9 points.
These results indicate that additional parameters alone do not explain the improvement.
While a future-state MLP operates on individual states and latent subtraction supplies a fixed, coordinate-wise change signal, IDM jointly processes ordered state pairs, preserving endpoint context and learning nonlinear, state-conditioned transition cues for planning refinement.

\begin{table}[htbp]
\caption{
Ablation of transition-aware future modeling on the NAVSIM navtest split.
Parameter-matched MLP variants have the same total parameter count as IDOL; all other components and training settings are fixed.
Parameter counts are reported in millions and rounded to one decimal place.
}
\label{tab:app_transition_ablation}
\centering
\small
\setlength{\tabcolsep}{4pt}
\renewcommand{\arraystretch}{1.05}
\resizebox{\linewidth}{!}{%
\begin{tabular}{l c c c c c c c}
\toprule
\textbf{Transition Modeling} & \textbf{Params (M)} & \textbf{NC} $\uparrow$ & \textbf{DAC} $\uparrow$ & \textbf{TTC} $\uparrow$ & \textbf{Comf.} $\uparrow$ & \textbf{EP} $\uparrow$ & \textbf{PDMS} $\uparrow$ \\
\midrule
Future State Only & 68.2 & 98.4 & 96.6 & 95.2 & \textbf{100} & 82.6 & \cellcolor{gray!20} 88.6 \\
Future State Only + Parameter-Matched MLP & 69.4 & 98.5 & 97.0 & 94.7 & \textbf{100} & 83.3 & \cellcolor{gray!20} 88.9 \\
Latent Difference & 68.2 & 98.6 & 96.8 & 95.7 & \textbf{100} & 82.8 & \cellcolor{gray!20} 89.0 \\
Latent Difference + Parameter-Matched MLP & 69.4 & \textbf{98.7} & 97.1 & \textbf{95.8} & \textbf{100} & 82.6 & \cellcolor{gray!20} 89.1 \\
IDM & 69.4 & \textbf{98.7} & \textbf{97.7} & 95.5 & \textbf{100} & \textbf{84.0} & \cellcolor{gray!20} \textbf{90.0} \\
\bottomrule
\end{tabular}%
}
\end{table}

\paragraph{Ablation on global dynamics fusion.}
We further study how to fuse the global dynamics feature produced by the inverse dynamics model.
In the inverse-dynamics-guided refinement network, the global dynamics feature provides holistic calibration for the query refined by spatial transition evidence.
We compare three simple fusion strategies: additive fusion, which directly injects the projected global feature into the query; concat-MLP fusion, which learns a nonlinear fusion from the concatenated query and global feature; and the MLN-based fusion used in IDOL.
As shown in Table~\ref{tab:app_fusion_ablation}, all variants obtain strong performance, indicating that the global inverse-dynamics cue is consistently beneficial for query refinement.
These results verify that IDOL's performance is not only due to using inverse-dynamics features, but also benefits from an effective query-calibration mechanism that better integrates global transition information into planning.

\begin{table}[htbp]
\caption{
Ablation of global dynamics fusion strategies in the inverse-dynamics-guided refinement network on the NAVSIM navtest split.
All variants use the same spatial branch and only replace the fusion operation between the refined ego query and the global dynamics feature.
}
\label{tab:app_fusion_ablation}
\centering
\small
\setlength{\tabcolsep}{6pt}
\renewcommand{\arraystretch}{1.05}
\begin{tabular}{l c c c c c c}
\toprule
\textbf{Fusion Strategy} & \textbf{NC} $\uparrow$ & \textbf{DAC} $\uparrow$ & \textbf{TTC} $\uparrow$ & \textbf{Comf.} $\uparrow$ & \textbf{EP} $\uparrow$ & \textbf{PDMS} $\uparrow$ \\
\midrule
Additive & \textbf{98.7} & 97.2 & \textbf{95.7} & \textbf{100} & 83.5 & \cellcolor{gray!20} 89.5 \\
Concat-MLP & \textbf{98.7} & 97.6 & 95.5 & \textbf{100} & 83.8 & \cellcolor{gray!20} 89.8 \\
MLN & \textbf{98.7} & \textbf{97.7} & 95.5 & \textbf{100} & \textbf{84.0} & \cellcolor{gray!20} \textbf{90.0} \\
\bottomrule
\end{tabular}
\end{table}

\paragraph{Sensitivity to anchor-set size.}
We evaluate IDOL with $K=128$, $K=256$, and $K=512$ trajectory anchors, keeping the anchor-construction procedure, backbone, world model, IDM, losses, and training settings unchanged.
As shown in Table~\ref{tab:app_anchor_ablation}, PDMS ranges from 89.6 to 90.0, indicating limited sensitivity to anchor-set size within the tested range.
A smaller vocabulary may limit trajectory coverage, whereas a denser vocabulary may introduce redundant candidates and increase ranking difficulty.
We retain $K=256$ as the default, as it achieves the highest PDMS while using fewer candidates than $K=512$.

\begin{table}[htbp]
\caption{
Sensitivity to trajectory-anchor vocabulary size on the NAVSIM navtest split.
Only the number of anchors is varied; the anchor-construction procedure and all other model and training settings remain unchanged.
}
\label{tab:app_anchor_ablation}
\centering
\small
\setlength{\tabcolsep}{6pt}
\renewcommand{\arraystretch}{1.05}
\begin{tabular}{c c c c c c c}
\toprule
\textbf{Number of Anchors} & \textbf{NC} $\uparrow$ & \textbf{DAC} $\uparrow$ & \textbf{TTC} $\uparrow$ & \textbf{Comf.} $\uparrow$ & \textbf{EP} $\uparrow$ & \textbf{PDMS} $\uparrow$ \\
\midrule
128 & \textbf{98.7} & 97.6 & \textbf{95.5} & \textbf{100} & 83.8 & \cellcolor{gray!20} 89.8 \\
256 & \textbf{98.7} & \textbf{97.7} & \textbf{95.5} & \textbf{100} & \textbf{84.0} & \cellcolor{gray!20} \textbf{90.0} \\
512 & \textbf{98.7} & 97.2 & 95.3 & \textbf{100} & \textbf{84.0} & \cellcolor{gray!20} 89.6 \\
\bottomrule
\end{tabular}
\end{table}

\section{Additional qualitative visualization}
\label{sec:Additional_qualitative}

\begin{figure*}[t]
  \centering
  \includegraphics[width=\textwidth]{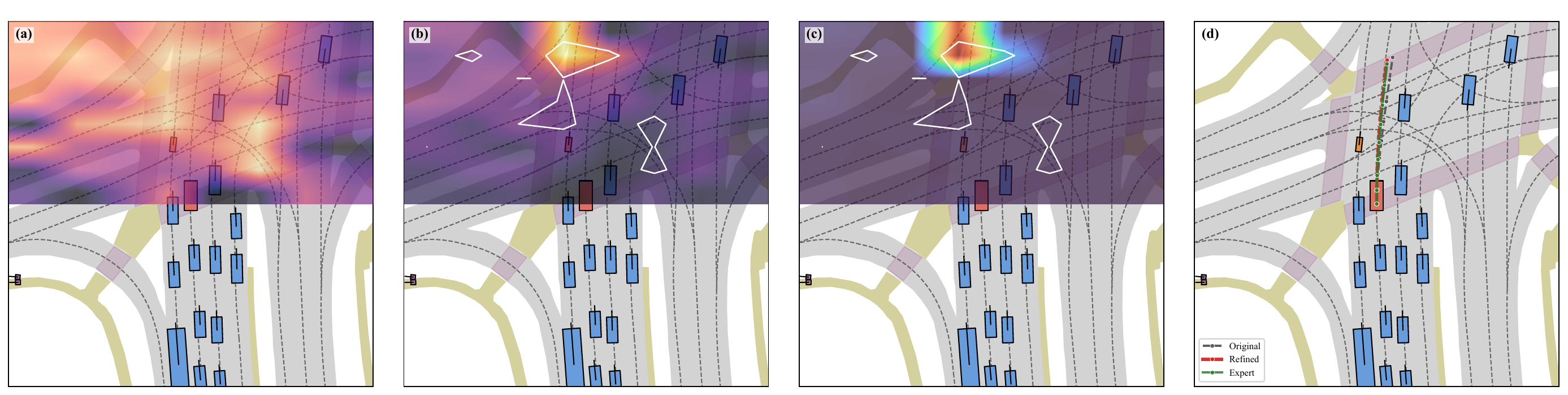}\par
  \vspace{0.4em}
  \includegraphics[width=\textwidth]{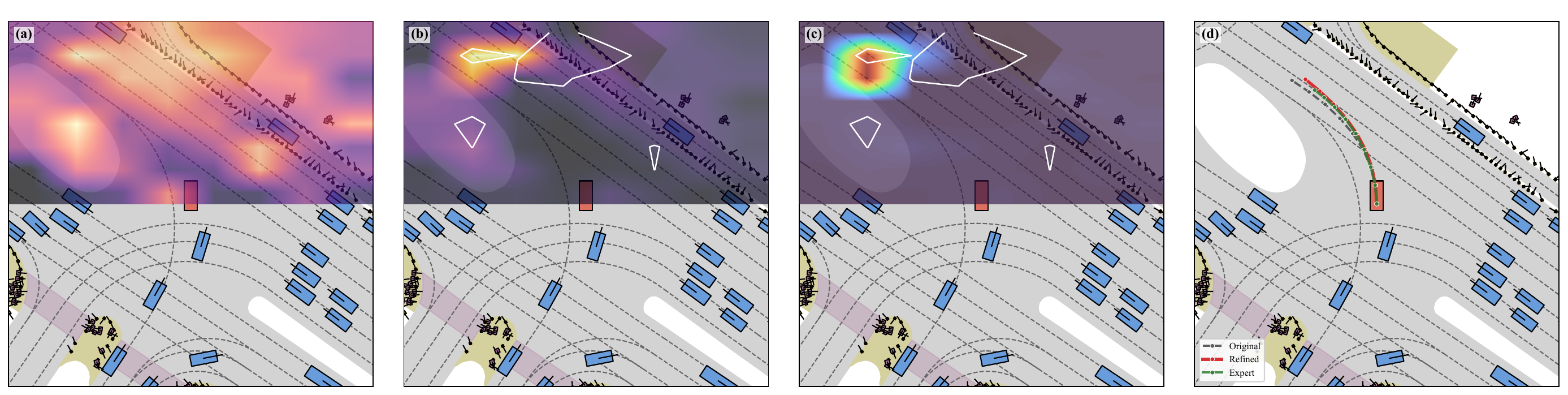}\par
  \vspace{0.4em}
  \includegraphics[width=\textwidth]{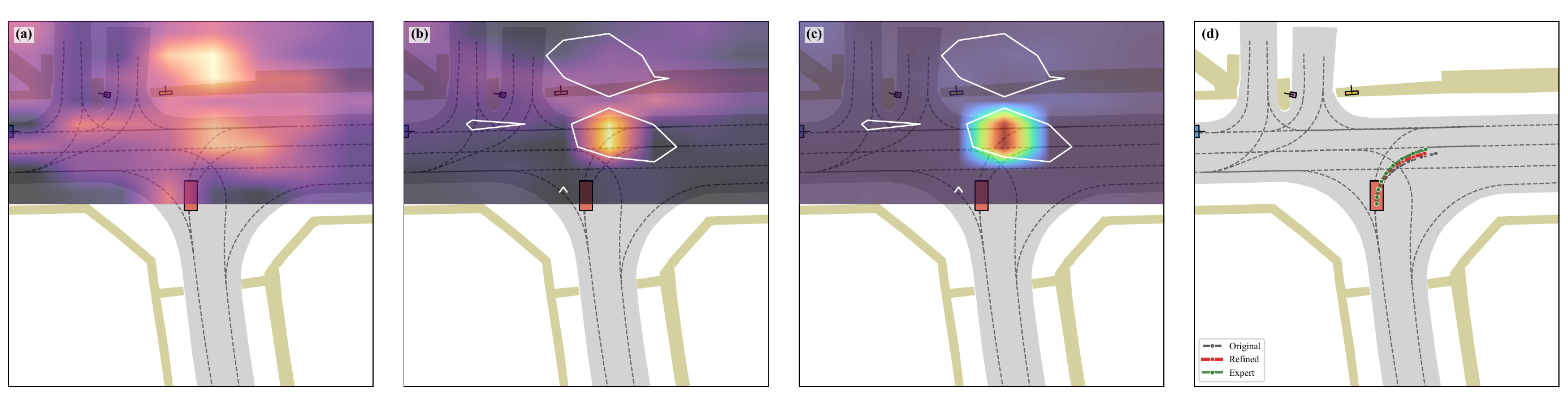}
  \caption{Qualitative visualization of inverse-dynamics refinement under three representative driving maneuvers on the NAVSIM navtest split. From top to bottom: straight driving, left turn, and right turn. In each row, panels (a)--(d) show the latent BEV change, the IDM dynamics inferred from adjacent BEV states, the IDM attention on dynamic queries, and the resulting refined trajectory, respectively.}
  \label{fig:appendix_idm_vis}
\end{figure*}

To further demonstrate the effectiveness of our proposed method, we expand the qualitative visualization experiments with additional representative driving scenarios.  Figure~\ref{fig:appendix_idm_vis} presents representative visualization results on the NAVSIM navtest split under three driving maneuvers, including straight driving, left turn, and right turn. Each example contains four panels that illustrate how the proposed IDM module extracts spatial dynamics from latent BEV transitions and uses them to refine the trajectory prediction.

Panel (a) shows the latent BEV change between two adjacent imagined BEV states produced by the world model. The highlighted regions indicate where the scene representation changes most significantly from frame 
$t$ to frame $t+1$
. Therefore, this panel serves as the reference evidence for spatially localized future dynamics, such as ego-motion-induced changes, nearby traffic participants, and lane-level scene evolution.

Panel (b) visualizes the spatial dynamics inferred by the IDM from the same adjacent BEV pair. Instead of directly displaying a semantic map, this panel shows the inverse-dynamics response over BEV queries. The white contour marks the high-change regions extracted from Panel (a), and is overlaid as a reference for comparing IDM responses with latent BEV transitions. The regions with stronger activation correspond to locations where IDM considers the transition between the two BEV states informative for explaining the underlying motion. The visual consistency between Panel (a) and Panel (b) suggests that IDM is not responding to arbitrary static background features, but is sensitive to the spatial regions where the latent scene actually evolves.

Panel (c) further shows the attention distribution from the trajectory query to the IDM spatial dynamics. While Panel (b) indicates where inverse-dynamics information is extracted, Panel (c) reveals which dynamic regions are actually attended to by the refinement query. The concentration of attention around the changing BEV regions demonstrates that the refinement process actively uses the IDM-derived spatial evidence, rather than treating IDM as an isolated auxiliary branch.

Panel (d) compares the original trajectory, the expert trajectory, and the refined trajectory. The original trajectory is decoded before applying the IDM-based closed-loop refinement, while the refined trajectory is decoded after the query has absorbed inverse-dynamics information through the IDM attention and fusion modules. Across straight, left-turn, and right-turn cases, the refined trajectory is visually better aligned with the expert trajectory, showing that the IDM-guided update helps correct the initial prediction toward a more plausible future motion.

Overall, these qualitative results provide direct visual evidence for the effectiveness of the IDM module. The latent BEV transition identifies where the future scene changes, the IDM spatial response recovers dynamics from those changing regions, and the query attention confirms that such information is used by the planner. The resulting trajectory refinement illustrates the practical value of this mechanism: IDM helps the model focus on motion-relevant spatial changes and improves the final trajectory prediction in different driving maneuvers. This supports our hypothesis that inverse-dynamics reasoning provides a meaningful bridge between latent world-model prediction and trajectory-level decision refinement.

\section{Failure Cases}
\label{sec:failure_cases}

\begin{figure}[t]
    \centering
    \includegraphics[width=\linewidth]{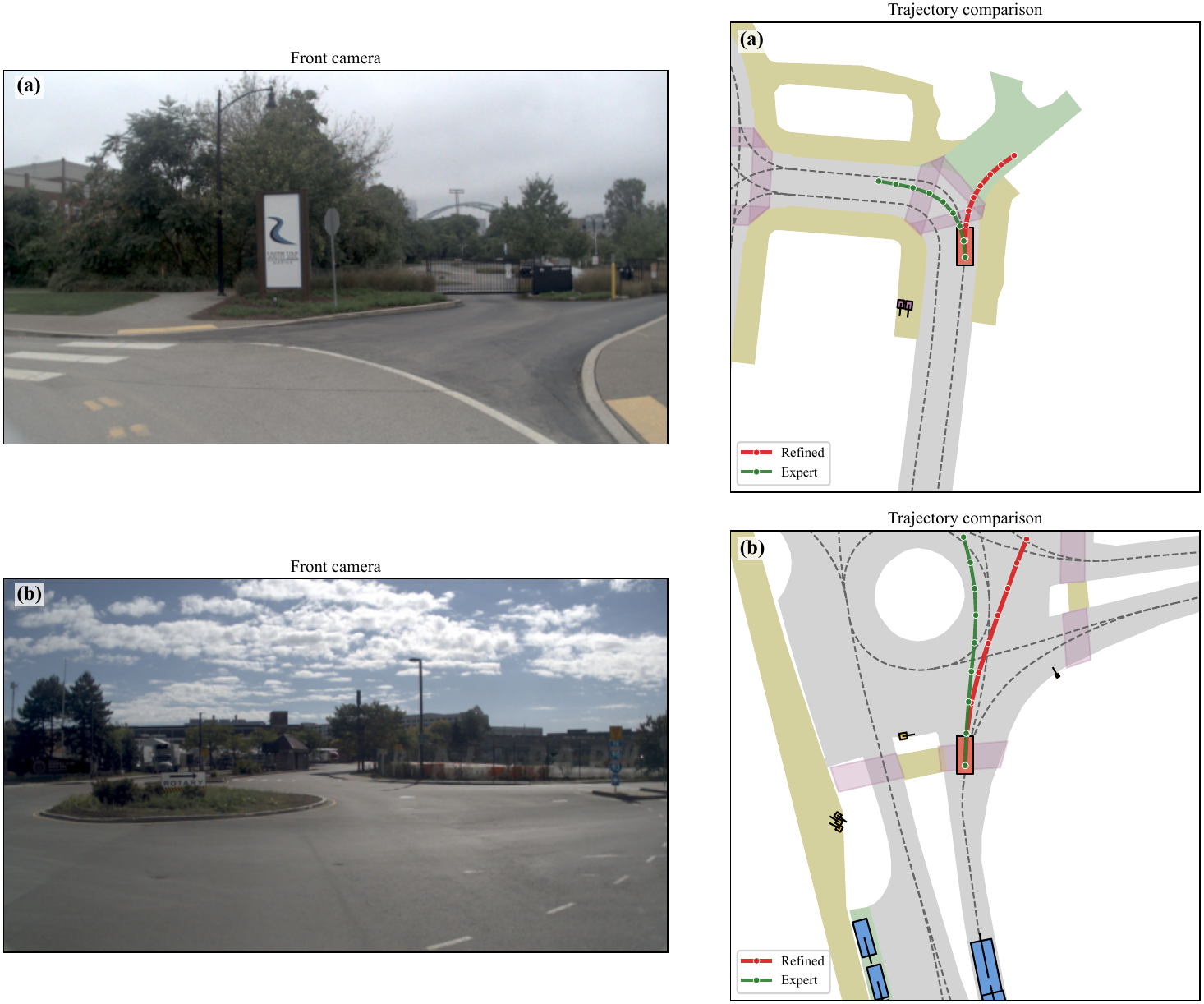}
    \caption{Representative failure cases on navtest.}
    \label{fig:failure_cases}
\end{figure}

We further analyze two representative failure cases on navtest, as shown in Figure~\ref{fig:failure_cases}. Each row contains the front-view camera image and the corresponding BEV trajectory comparison. The refined trajectory denotes the final prediction after IDM-based refinement, and the expert trajectory is used as the reference.

In the first case, the ego vehicle approaches a turning area with ambiguous road geometry. The refined trajectory deviates clearly from the expert trajectory and follows an incorrect spatial direction. This indicates that when the model fails to infer the correct maneuver intention, IDM refinement alone may not fully recover the desired path.

In the second case, the scene involves a complex intersection or roundabout-like structure. Although the refined trajectory captures part of the motion trend, it remains spatially offset from the expert trajectory. This suggests that topology-sensitive scenarios still require more precise reasoning about lane connectivity, road curvature, and maneuver timing.

These cases show that IDM provides useful dynamic cues, but its effectiveness depends on whether such cues can be correctly fused with scene topology and trajectory intent. Future work may improve these limitations with stronger topology-aware constraints and uncertainty-aware refinement.

\section{Limitations and future work}
\label{sec:limitations_future_work}

Although IDOL shows strong performance across NAVSIM benchmarks, several limitations remain.

\paragraph{Model capacity.}
Our inverse dynamics model is designed to be lightweight, which keeps the framework efficient but may limit its ability to capture more complex long-horizon interactions and subtle motion patterns. Future work could scale the IDM with larger temporal contexts, stronger transition modeling, or pretraining on broader driving sequences.

\paragraph{Dependence on latent future prediction.}
IDOL relies on the quality of the imagined latent BEV futures produced by the world model. When future prediction is inaccurate in highly ambiguous or rare scenarios, the decoded inverse-dynamics cues may also become less reliable. Improving the robustness of latent world modeling is therefore an important direction.

\paragraph{Fixed refinement strategy.}
The current closed-loop refinement uses a fixed refinement setting, while different scenes may require different levels of future-aware reasoning. Adaptive refinement strategies could further improve stability and avoid unnecessary iterative updates.

\paragraph{Broader evaluation.}
Although our experiments cover both standard and challenging NAVSIM settings, future work will further evaluate inverse-dynamics-guided planning in more interactive, long-tail, and real-world driving scenarios.

\newpage

\end{document}